\documentclass{article}

\usepackage{amsmath}
\usepackage{amssymb}
\usepackage{booktabs}
\usepackage{graphicx}
\usepackage{subcaption}
\usepackage{enumitem}
\usepackage{multirow}
\graphicspath{{images/}}
\usepackage[final]{corl_2026} 
\makeatletter
\renewcommand{\@conferencelocation}{Austin, Texas, USA}
\makeatother

\title{ContactGuard: Pre-Contact Execution Monitoring with Action-Conditioned Latent World Models}

\newcommand{\twoline}[2]{\begin{tabular}[b]{@{}c@{}}#1\\#2\end{tabular}}

\hypersetup{
  pdftitle={ContactGuard: Pre-Contact Execution Monitoring with Action-Conditioned Latent World Models},
  pdfauthor={Gehan Zheng, Matthew Johnson-Roberson, Weiming Zhi}}

\author{Gehan Zheng$^{1}$,
Matthew Johnson-Roberson$^{1}$,
Weiming Zhi$^{2, 1}$}

\begin{document}
\maketitle


\begin{abstract}
Contact-rich manipulation failures are often detected only after the robot has committed to contact. This is especially limiting in wrist-camera setups: close gripper--object views help observe contact, but a poor approach may already push, miss, slip, or disturb the object before conventional detectors react. We introduce \emph{ContactGuard}, a pre-contact execution monitor for chunked visuomotor policies. Given the policy's planned action chunk, ContactGuard predicts its short-horizon consequence in latent visual space and aborts if the predicted future latent indicates likely failure. Its latent world model is trained from unlabelled robot trajectories to predict compact multi-view visual embeddings under planned actions, avoiding pixel-level video prediction. A lightweight failure probe is then trained from a small labelled set of pre-contact clips. At deployment, ContactGuard anchors prediction before an imminent contact event, rolls the model forward under the policy's own actions, and verifies the predicted post-contact latent. Across real-world contact-rich manipulation tasks, ContactGuard consistently benefits from action-conditioned latent prediction over current-state and direct-prediction alternatives. Counterfactual action perturbations further show that its failure score responds to the specific action being proposed. We deploy the same frozen monitor online with the visuomotor policy, where it provides a pre-contact abort signal without modifying the underlying policy.
\end{abstract}

\keywords{Imitation learning, Latent world models for robot learning
}
\footnotetext[1]{College of Connected Computing, Vanderbilt University, USA.}
\footnotetext[2]{School of Computer Science, The University of Sydney, Australia.}
\section{Introduction}
\label{sec:intro}

Reliable manipulation requires more than visually plausible motion. In contact-rich tasks, small errors near contact can determine success or failure: the gripper may approach off-centre, nudge the object away, pinch only an edge, miss a deformable region, or close before the object is seated. These failures are difficult for visuomotor policies to avoid under occlusion, clutter, deformable objects, and distribution shift, and they are harder to correct once contact has disturbed the scene. This motivates \emph{pre-contact execution monitoring}: instead of detecting whether the robot has already failed, the monitor asks whether the action it is about to execute is likely to fail. Chunked visuomotor policies make this question natural, because an action chunk often contains a meaningful interaction event, such as gripper closure or object contact. If the robot can evaluate the consequence of that chunk before contact, it can stop a poor interaction before committing to it.

A natural way to reason about future consequences is to learn a world model. For manipulation monitoring, however, the model need not generate photorealistic future frames. It only needs a representation that preserves outcome-relevant information: whether the planned interaction is likely to succeed or fail. We therefore use action-conditioned latent prediction, which encodes multi-view observations into compact visual embeddings and learns how those embeddings evolve under robot actions. This follows the Joint-Embedding Predictive Architecture (JEPA) principle~\citep{assran_ijepa_2023}, where a predictor is trained to forecast the embedding of a target signal rather than reconstruct it in pixel space; predicting directly in representation space avoids the cost and ambiguity of pixel-level video generation. We introduce \emph{ContactGuard}, a pre-contact monitor built on this principle. A latent world model is trained from unlabelled robot trajectories using next-latent supervision; after training, its encoder and predictor are frozen. A lightweight logistic-regression probe is then trained on labelled pre-contact clips using only the predicted future latent, separating task-agnostic visual dynamics learning from small-data outcome supervision.

\begin{figure}[t]
    \centering
    \includegraphics[width=0.90\linewidth]{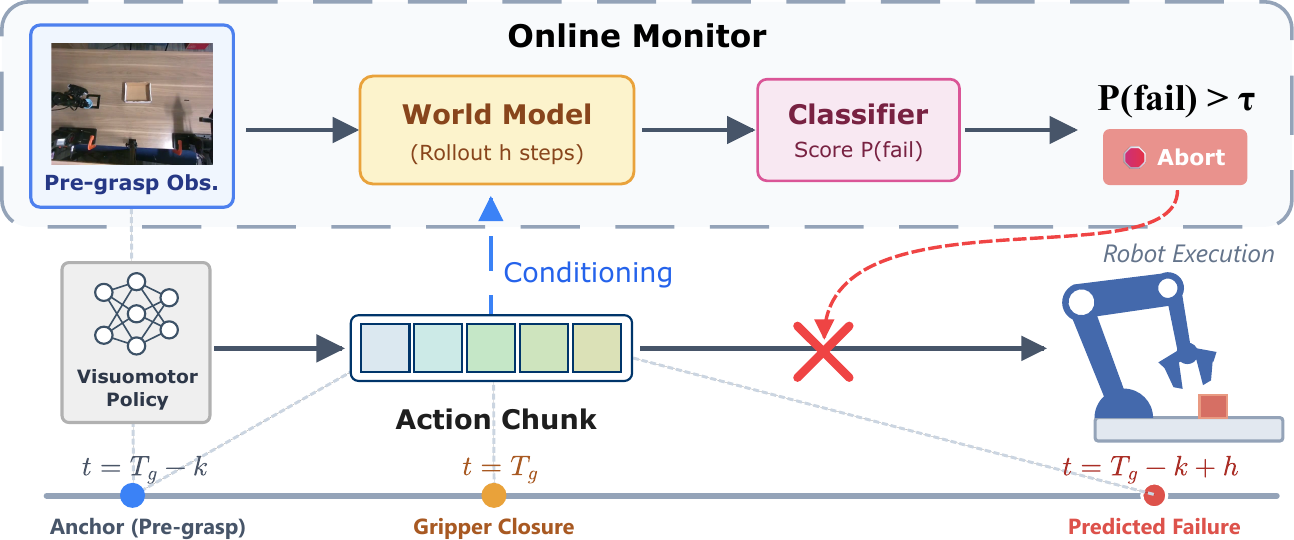}
    \caption{Pre-contact online monitoring of the policy's action chunk.
    Anchored $k$ steps before planned gripper closure $T_g$, the frozen world model rolls forward $h$ steps under the \emph{same} proposed actions. A lightweight classifier scores the predicted post-contact latent at $T_g{-}k{+}h$. If $P(\mathrm{fail})>\tau$, ContactGuard aborts at $T_g{-}k$, before contact, to prevent the predicted failure.}
    \label{fig:teaser}
\end{figure}

At deployment, ContactGuard runs alongside an existing chunked visuomotor policy as a \emph{policy-decoupled predictive verifier}. The deployed policy is treated as a black-box proposer: ContactGuard consumes only the current observation and the concrete action chunk already emitted by the policy, without accessing policy internals or requiring joint training. It scans the upcoming chunk for an imminent contact event, anchors prediction shortly before that event, rolls the independently trained latent world model forward under the proposed actions, and applies the frozen probe to the predicted post-contact latent. The system aborts before contact if that specific chunk is likely to lead to failure; otherwise the policy continues unchanged.

Our experiments test whether this action-conditioned future latent contains failure information not readily available from the current observation or planned action alone. Across real-world contact-rich manipulation tasks, predicted future latents improve failure prediction over current-state and direct-prediction baselines, while counterfactual action perturbations show that the predicted failure score depends on the specific proposed action. These results suggest that latent world models can serve not only as planning or representation-learning modules, but also as practical execution monitors for preventing likely failures before physical contact.

Our contributions are:
\begin{itemize}[leftmargin=*, align=left]
    \item We formulate pre-contact execution monitoring for chunked visuomotor policies, where the robot evaluates the likely consequence of an imminent contact action before committing.
    \item We introduce \emph{ContactGuard}, an action-conditioned latent world-model monitor that rolls out compact multi-view visual embeddings under the policy's planned action chunk.
    \item We show that predicted future latents provide failure information beyond current observations and raw planned actions, through matched current-state and direct-prediction baselines together with counterfactual action perturbations.
    \item We demonstrate live real-robot pre-contact aborts without candidate action search, pixel-level video prediction, or modification of the underlying visuomotor policy.
\end{itemize}


\section{Related Work}
\label{sec:related}

\begin{figure}[t]
\centering
\includegraphics[width=\linewidth]{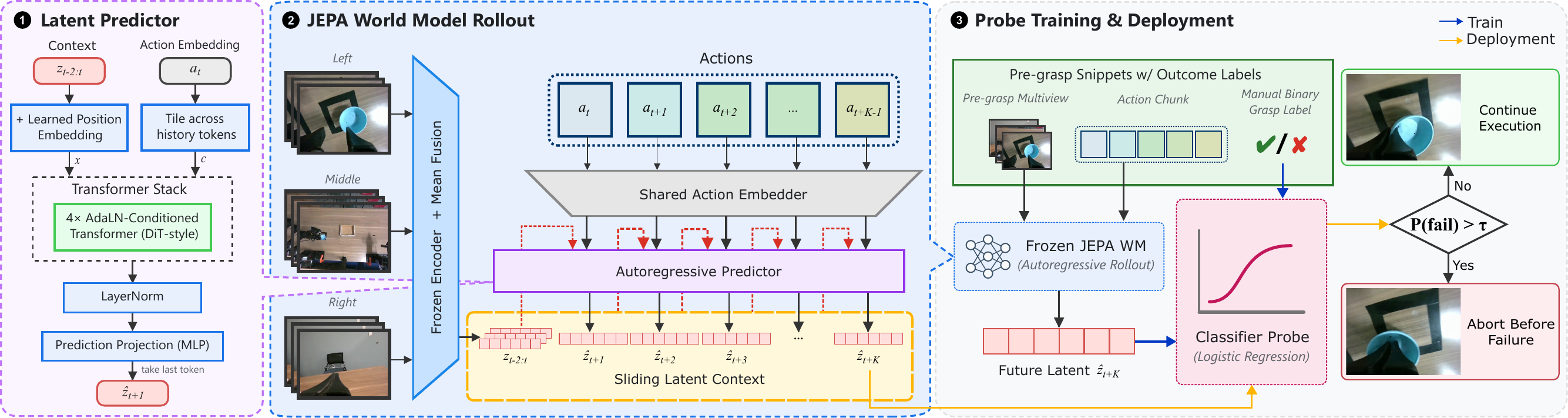}
\caption{ContactGuard overview.
A frozen encoder $E_\phi$, shared across views, maps camera observations to a latent context. An action-conditioned predictor rolls this context forward under the policy's planned action chunk to produce future latents $\hat z_{t+1:t+K}$. A lightweight probe is trained offline on labelled pre-contact clips and used online to score the predicted post-contact latent $\hat z_{t+K}$. If the predicted failure probability exceeds threshold $\tau$, execution is aborted before contact. Training combines teacher-forced next-latent prediction with the autoregressive rollout used at inference.}
\label{fig:overview}
\end{figure}

\textbf{Latent world models for robot manipulation:} Learning compact dynamics from pixels is central to model-based RL~\citep{ha_world_2018,hafner_learning_2019}. Dreamer~\citep{hafner_dream_2020,hafner_mastering_2024} imagines trajectories in latent space for policy learning, while TD-MPC~\citep{hansen_temporal_2022} learns task-oriented latent dynamics for model-predictive control. Latent forward models have also been used for visuomotor trajectory optimisation~\citep{srinivas_universal_2018}, and action-conditioned video prediction for model-predictive manipulation~\citep{finn_deep_2017}. Recent learned simulators and generative world models extend this direction to richer visual and interaction rollouts for planning, policy learning, and data generation~\citep{zhu_irasim_2025,barcellona_dream_2025}. Most closely related is LeWorldModel~\citep{maes_leworldmodel_2026}, which trains an end-to-end latent world model with next-embedding prediction and an anti-collapse Gaussian regulariser. We adopt this latent-prediction perspective, but use it for pre-contact failure monitoring rather than planning or policy learning. In contact-rich manipulation, small changes around contact can decide whether an object is nudged, slips, or remains inside the gripper. We therefore ask whether action-conditioned future latents preserve enough outcome information to predict failure before contact occurs. 

\textbf{Failure prediction and execution monitoring:}
Robust robot deployment has motivated visual and vision-language detectors for manipulation success and failure~\citep{du_vision-language_2023,duan_aha_2024}, as well as robotic reward models learned from comparisons or large reward datasets~\citep{liang_robometer_2026,lee_roboreward_2026}. Other methods monitor trajectory rarity \cite{DOSE3}, or monitor learned policies during execution: FIPER predicts failures of generative robot policies, while FAIL-Detect detects runtime distribution shift from successful demonstrations~\citep{romer_failure_2025,xu2025faildetect}; RND provides another success-only novelty signal~\citep{burda2019rnd}, and SAFE learns a supervised failure detector across manipulation tasks~\citep{gu2025safe}. Out-of-distribution metrics for visuomotor policies, such as Rewind-IL~\cite{zheng2026rewind} and PATCH~\cite{zhou2026patch}, also enable accurate failure prediction. Closely related, SIRIUS jointly trains its policy and latent dynamics model in a shared latent space and alternates them to simulate future policy rollouts for runtime monitoring~\citep{liu2024sirius}. Sirius-Fleet similarly uses visual world-model predictions for failure monitoring in multi-task interactive robot learning~\citep{liu2024siriusfleet}. Learned rewards have also been used to score imagined world-model rollouts for model-based policy learning~\citep{amigo2026coupled}, and world-model latent predictions for failure classification and anomaly detection in autonomous inspection~\citep{ho2026world}; ContactGuard instead performs event-triggered, pre-contact gating of a single proposed action chunk, scoring its failure probability rather than reward. We therefore do not claim imagined future-state monitoring itself as novel. ContactGuard addresses a different deployment interface: the underlying policy remains an external black-box proposer, while an independently trained world model verifies the \emph{concrete action chunk already proposed by that policy} at a pre-contact commitment point and may veto it before scene-changing contact. This decouples verification from policy training, so the monitor attaches to an otherwise unchanged visuomotor policy. For grasping, prior work has studied early failure prediction with sequence models and interactive visual predictors, where the robot probes or partially executes an action before completing a pick~\citep{damak_robot_2025}. ContactGuard instead predicts failure before contact by rolling the proposed chunk from a pre-contact observation to a predicted post-contact latent.

\textbf{Chunked visuomotor policies and test-time verification:}
Imitation learning~\cite{Diff_templates,ravichandar2020recent,periodic} enables motion generation without requiring structured motion planning~\cite{GeoFab_gloabL_opt}. Action chunks are a natural unit for short-horizon visuomotor prediction. ACT~\citep{zhao_learning_2023} predicts action sequences with a transformer policy, while Diffusion Policy~\citep{chi_diffusion_2024} generates chunks through conditional denoising. These chunks often contain meaningful interaction events such as approach, gripper closure, and lift, making them suitable for consequence prediction before execution, especially over long horizons~\cite{li2026tripilot}. Recent methods use learned rewards or world models to score multiple candidate actions for test-time verification~\citep{dai_rover_2025} or post-training~\citep{sun_atomvla_2026, tang2026autointervene}. ContactGuard addresses a different setting: it does not search over candidate chunks or modify the action generator. It monitors the single chunk proposed by an existing policy and uses the predicted future latent as a pre-contact failure signal.

\section{Latent World Models for Pre-Contact Grasp Monitoring}
\label{sec:method}

We present \emph{ContactGuard}, a pre-contact grasp monitor that predicts whether an imminent grasp is likely to fail before the gripper closes. It combines a JEPA-style latent world model with a frozen linear failure probe. The world model is trained from unlabelled robot trajectories to predict action-conditioned future latents; the probe is trained from a smaller labelled grasp set to score the predicted post-contact latent. At deployment, the frozen model and probe run alongside a visuomotor policy and abort execution when the planned action chunk is predicted to fail.

We first define the world model and probe data, then describe the LeWM-style latent predictor, its multi-view architecture, the failure probe, and the online pre-contact monitor.

\subsection{Problem Setting and Data}

ContactGuard separates unlabelled latent dynamics learning from small-data outcome supervision. The world-model dataset contains robot trajectories
\begin{equation}
    \mathcal{D}_{\mathrm{wm}}
    =
    \left\{
    \left(o_{1:T_n}^{1:V}, a_{1:T_n}\right)_n
    \right\}_{n=1}^{N},
\end{equation}
where $o_t^{1:V}$ are synchronised observations from $V$ fixed cameras and $a_t \in \mathbb{R}^d$ is the joint-space action. No grasp-success labels are used to train the world model.

The probe dataset $\mathcal{D}_{\mathrm{probe}}$ contains labelled pre-contact grasp clips. Each clip includes a short multi-view history $o_{t-C+1:t}^{1:V}$, the planned action chunk $a_{t:t+K-1}$, and a binary label $y \in \{0,1\}$, where $y{=}1$ denotes failure. After the world model is frozen, the probe learns to map the predicted future latent to failure likelihood.

\subsection{Background: LeWM-style Latent Prediction}
\label{sec:prelim}

Figure~\ref{fig:overview} shows the ContactGuard architecture; at its core lies a latent JEPA model~\citep{maes_leworldmodel_2026}. A visual encoder $E_\phi$ maps an observation to a latent embedding $z_t = E_\phi(o_t) \in \mathbb{R}^D$, and an action-conditioned causal predictor $P_\theta$ predicts the next latent from a length-$C$ context:
\begin{equation}
    \hat{z}_{t+1}
    =
    P_\theta(z_{t-C+1:t}, a_t).
\end{equation}
The predictor is a causal Transformer conditioned on actions through AdaLN-zero modulation. The original LeWorldModel formulation combines next-embedding regression with the SIGReg anti-collapse regulariser~\citep{maes_leworldmodel_2026}, which encourages the latent distribution to match a standard Gaussian on random projections. We reuse the encoder--predictor interface and further adapt the interface to multi-view observations, action-aligned prediction windows, and downstream pre-contact failure readout.

\subsection{Multi-View Latent World Model}
\label{sec:arch}

 At each time step, the $V$ camera observations are encoded independently by a shared frozen DINOv2-S encoder~\citep{oquab_dinov2_2024} (CLS token), which receives no task-specific training. The per-view embeddings are mean-pooled and passed through a fixed linear projection, frozen with the encoder, to produce a single latent $z_t \in \mathbb{R}^D$. This keeps the latent dimension fixed as the number of cameras changes, while allowing informative views to compensate for partial occlusions. A shared action embedder maps each joint-space action $a_t \in \mathbb{R}^d$ into the same $D$-dimensional space. The predictor consists of four AdaLN-zero-conditioned Transformer blocks followed by a two-layer prediction MLP. We denote the full Transformer-plus-projection module by $P_\theta$. During training, $P_\theta$ is supervised with teacher-forced one-step transitions over a context window of $C$ frames and $L$ rollout targets per sample. At deployment, the same predictor is rolled out autoregressively for $K$ steps under the planned action chunk.

We train $P_\theta$ with the same action-conditioned one-step interface used during rollout. Let $z_0,\ldots,z_{C+L-1}$ be the latents in a training window and let $u_i$ be the action that drives the transition from frame $i{+}C{-}1$ to frame $i{+}C$. For each supervised step $i \in \{0,\ldots,L{-}1\}$, the predictor receives a length-$C$ real-latent context and the aligned action:
\begin{equation}
\label{eq:supervision}
    \hat z_{i+C}
    =
    P_\theta\bigl(z_{i:i+C-1}, u_i\bigr).
\end{equation}
The objective is whitened next-latent regression, applied to the teacher-forced predictions and to the autoregressive rollout $\hat z^{\mathrm{ro}}$ at a sparse set of horizons $\mathcal{H}$ up to the deployment horizon:
\begin{equation}
\label{eq:loss}
    \mathcal{L}
    =
    \frac{1}{m}\sum_{i=0}^{m-1}
    \bigl\|\Sigma^{-\frac12}(\hat z_{i+C} - z_{i+C})\bigr\|_2^2
    +
    \frac{1}{|\mathcal{H}|}\sum_{h \in \mathcal{H}}
    \bigl\|\Sigma^{-\frac12}(\hat z^{\mathrm{ro}}_{C+h-1} - z_{C+h-1})\bigr\|_2^2 ,
\end{equation}
where $\Sigma$ is the per-dimension variance of the frozen encoder's latents. Thus each training target matches one inference-time rollout step: teacher-forced real latents in the sliding context, plus the same autoregressive rollout used at deployment.

\subsection{Linear Failure Probe on Predicted Latents}
\label{sec:probe}

The world model is task-agnostic. After training, we freeze $E_\phi$ and $P_\theta$ and use the $K$-step predicted latent as the feature for a lightweight failure readout. For each labelled clip, we anchor the rollout at $t = T_g - k_{\mathrm{pre}}$, unroll the frozen predictor for $K$ steps under the recorded action chunk, and use $\hat z_{t+K}$ as the probe input. We set $K > k_{\mathrm{pre}}$ so that the readout lies shortly after the planned gripper closure, where success or failure is more likely to be expressed in the latent.

Each feature is standardised using training-split statistics, $\tilde z_{t+K} = (\hat z_{t+K} - \mu)/\sigma$, and scored by a linear logistic probe:
\begin{equation}
\label{eq:probe}
    P(\mathrm{fail}\mid \hat z_{t+K})
    =
    \sigma\!\left(w^\top \tilde z_{t+K} + b\right).
\end{equation}
The probe parameters are fit on the training data with an $\ell_2$-penalised, class-balanced logistic loss:
\begin{equation}
\label{eq:probe_loss}
    \min_{w,b}
    \;
    \frac{1}{2}\|w\|_2^2
    +
    \rho
    \sum_{i=1}^{N}
    s_{y_i}
    \log\!\left(
    1 + \exp\!\left(-(2y_i-1)(w^\top \tilde z_i + b)\right)
    \right),
    \qquad
    s_c = \frac{N}{2N_c}.
\end{equation}
Here $N_c$ is the number of samples from class $c$, and the class weights compensate for success/failure imbalance. For each task, the fitted $(w,b)$ are frozen and used for all test-set and online evaluations.

\subsection{Online Pre-Contact Grasp Monitor}
\label{sec:monitor}

The offline failure score becomes an online gate evaluated before a task-defined imminent contact event. ContactGuard separates triggering from evaluation: a lightweight, task-specific trigger decides when the monitor should run, while the action-conditioned latent predictor decides whether the planned chunk should continue or be vetoed. In this paper, we instantiate the trigger for grasp closure using the commanded gripper open-to-close transition, a policy-agnostic cue that requires no learning; designing general-purpose triggers for other contact events is orthogonal to our contribution. The monitor runs alongside a chunked visuomotor policy such as ACT~\citep{zhao_learning_2023} and maintains the last $C$ multi-view observations and upcoming actions. Each control step follows four stages.

\textbf{Trigger:} Scan the upcoming chunk for the first open-to-close gripper transition, and denote its offset by $g$. If no closure is found, continue execution. \textbf{Anchor:} Activate the monitor when the planned closure is at most $k_{\mathrm{pre}}$ frames ahead, matching the pre-contact offset used for probe training. \textbf{Rollout and score:} Encode the observation history with $E_\phi$, roll out $P_\theta$ for $K$ steps under the planned actions, and apply the frozen probe to $\hat z_{t+K}$. \textbf{Abort decision:} If $P(\mathrm{fail}) > \tau$, abort before the gripper closes; otherwise continue the policy unchanged. The threshold $\tau$ is selected once per task on the validation split and then frozen for test-set and real-robot evaluations. Because the policy executes action chunks, the monitor runs within the chunk execution window and does not stall control unless an abort is issued.

\section{Experimental Results}
\label{sec:result}

We instantiate ContactGuard on grasp closure, a canonical short-horizon imminent-contact event with challenging visual geometry, occlusion, object variation, and a strict real-time pre-contact budget. We evaluate four grasp settings with ACT~\citep{zhao_learning_2023} as the underlying chunked visuomotor policy. All metrics evaluate the predictor's decision at the triggered pre-contact state, not the trigger itself or post-abort recovery. Our experiments ask four questions:
\begin{enumerate}[leftmargin=*, align=left]
\item \textbf{Prediction quality.} Can predicted future latents anticipate execution outcomes better than single-view world models, current-state monitoring, and established runtime failure detectors?
\item \textbf{Information source.} Does the signal arise from the imagined consequence of the \emph{specific proposed action}, rather than from the current observation, visual shortcuts, or generic motion cues?
\item \textbf{Robot-state fusion.} Does adding proprioceptive state improve grasp-failure prediction?
\item \textbf{Runtime.} Is the forecast fast enough for the pre-contact window before gripper closure?
\end{enumerate}

\begin{figure}[t]
\centering
\begin{subfigure}[t]{0.48\linewidth}
    \centering
    \includegraphics[width=\linewidth]{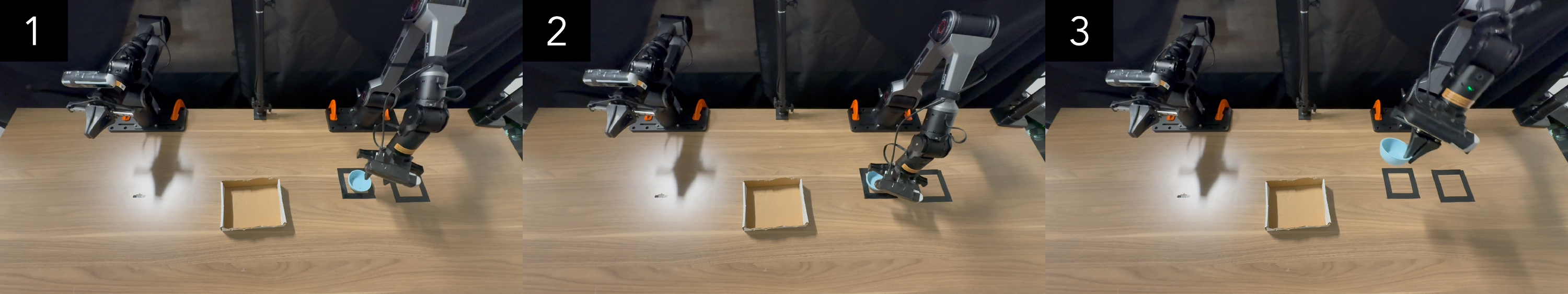}
    \caption{Pick-and-place (cup)}
\end{subfigure}
\begin{subfigure}[t]{0.48\linewidth}
    \centering
    \includegraphics[width=\linewidth]{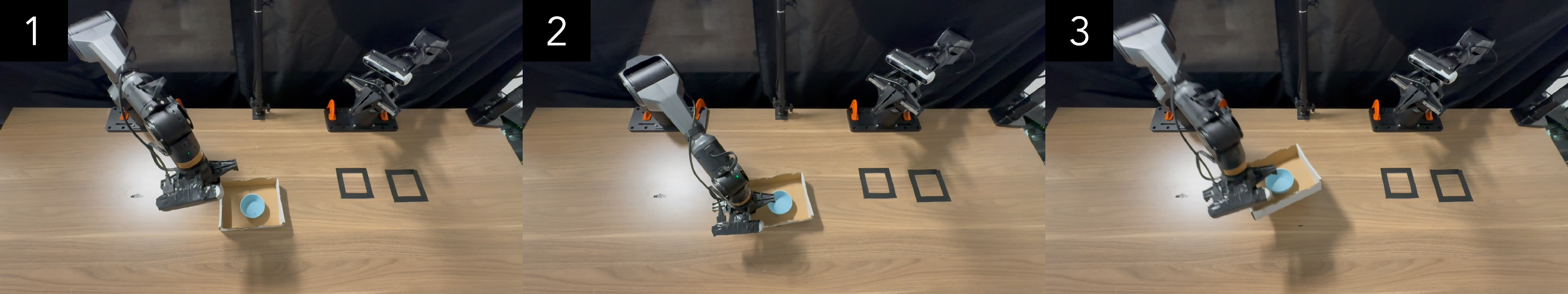}
    \caption{Pick-and-place (box)}
\end{subfigure}
\begin{subfigure}[t]{0.48\linewidth}
    \centering
    \includegraphics[width=\linewidth]{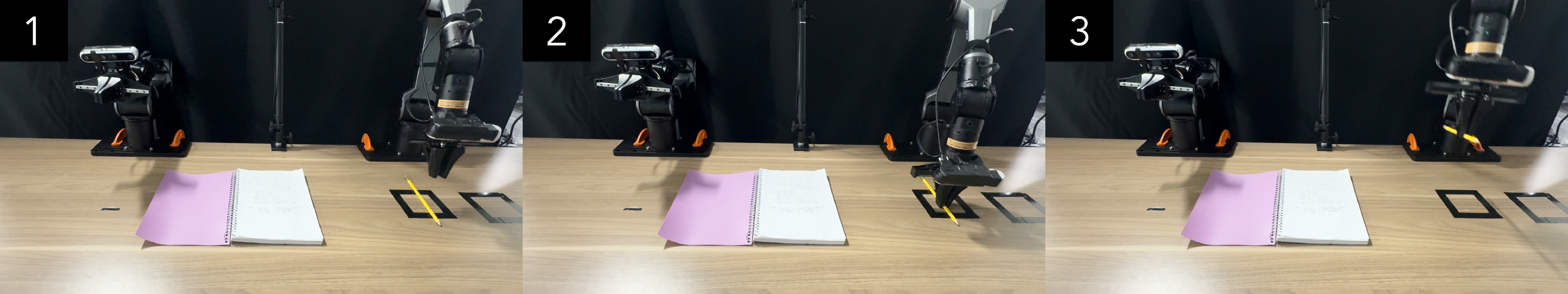}
    \caption{Pencil-and-notebook}
\end{subfigure}
\begin{subfigure}[t]{0.48\linewidth}
    \centering
    \includegraphics[width=\linewidth]{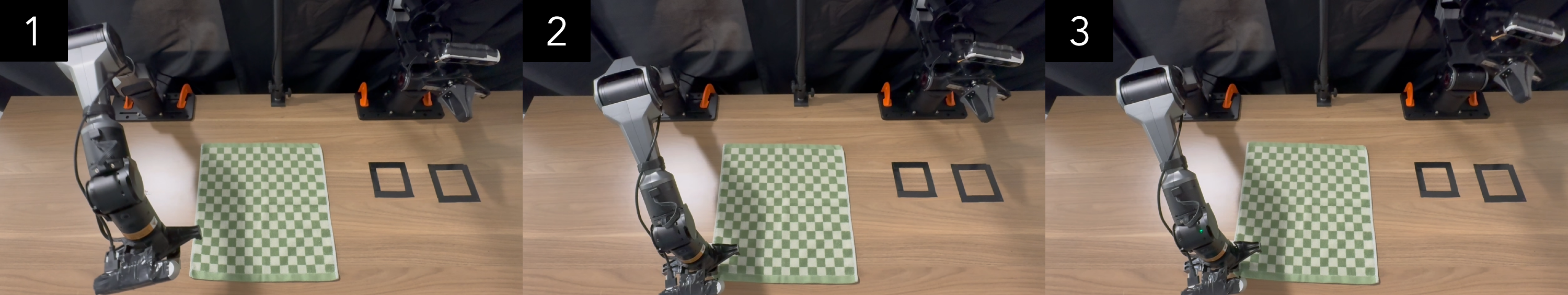}
    \caption{Towel-fold}
\end{subfigure}
\caption{Evaluated tasks. Each panel shows a representative successful rollout from the middle camera, with columns showing the pre-grasp, grasp, and post-grasp phases.}
\label{fig:tasks_setup}
\end{figure}

\subsection{Experimental Setup}

\textbf{Robot, tasks, and data:}
All experiments use a 14-DoF AgileX Piper dual-arm robot with three synchronized RGB cameras. We evaluate four labelled grasp settings: cup and box grasps in pick-and-place, pencil grasping in pencil-and-notebook, and towel grasping in towel-fold. These settings require precise end-effector--object alignment near contact, where small approach errors cause missed grasps, slips, or object disturbance. The world model is trained on unlabelled real-robot trajectories from ACT rollouts and human teleoperation. For downstream evaluation, we extract grasp attempts, assign binary success/failure labels, and define $T_g$ as the first gripper-command frame crossing the closure threshold. Each labelled set contains roughly 250 attempts; the live test episodes are collected separately and excluded from world-model training, probe fitting, threshold selection, and ablation tuning.

\begin{table}[t]
\centering
\begin{minipage}[t]{0.56\textwidth}
\vspace{0pt}
\small
\caption{Closed-loop real-robot results. We report precision, recall, false-abort rate (FAR$={}$FP/(FP+TN); lower is better), ROC AUC, and balanced accuracy over realized $P(\mathrm{fail})$ scores from $N{=}50$ monitored live rollouts per grasp setting. Best results in bold.}
\label{tab:realrobot}
\end{minipage}\hfill
\begin{minipage}[t]{0.42\textwidth}
\vspace{0pt}
\small
\caption{Offline ROC AUC on the same saved rollouts. Task-specific probes/detectors fit the training split only; world models use unlabeled pretraining. Baseline inputs are defined in the text.}
\label{tab:external_baselines}
\end{minipage}
\par\medskip
\begin{minipage}[b]{0.56\textwidth}
\centering
\small
\setlength{\tabcolsep}{2pt}
\begin{tabular}[b]{@{}llccccc@{}}
\toprule
Task & Method & Prec. & Rec. & FAR & AUC & BAcc \\
\midrule
\multirow{3}{*}{Cup}
 & Direct-linear & 0.810 & 0.680 & 0.160 & 0.875 & 0.760 \\
 & Single view  & 0.875 & 0.840 & 0.120 & 0.930 & 0.860 \\
 & Ours & \textbf{0.917} & \textbf{0.880} & \textbf{0.080} & \textbf{0.960} & \textbf{0.900} \\
\multirow{3}{*}{Box}
 & Direct-linear & 0.656 & 0.955 & 0.393 & 0.963 & 0.781 \\
 & Single view  & 0.750 & 0.955 & \textbf{0.250} & 0.925 & 0.852 \\
 & Ours & \textbf{0.759} & \textbf{1.000} & \textbf{0.250} & \textbf{0.966} & \textbf{0.875} \\
\multirow{3}{*}{Pencil}
 & Direct-linear & 0.789 & 0.682 & 0.143 & 0.833 & 0.769 \\
 & Single view   & \textbf{0.938} & 0.682 & \textbf{0.036} & 0.907 & 0.823 \\
 & Ours & 0.864 & 0.864 & 0.107 & \textbf{0.930} & \textbf{0.878} \\
\multirow{3}{*}{Towel}
 & Direct-linear & 0.929 & 0.520 & \textbf{0.040} & 0.898 & 0.740 \\
 & Single view   & 0.941 & 0.640 & \textbf{0.040} & 0.890 & 0.800 \\
 & Ours & \textbf{0.950} & \textbf{0.760} & \textbf{0.040} & \textbf{0.904} & \textbf{0.860} \\
\bottomrule\end{tabular}
\end{minipage}\hfill
\begin{minipage}[b]{0.42\textwidth}
\centering
\small
\renewcommand{\arraystretch}{1.39}
\setlength{\tabcolsep}{2pt}
\begin{tabular}[b]{@{}lcccc@{}}
\toprule
Detector & Cup & Box & Pencil & Towel \\
\midrule
FAIL-Detect & 0.840 & 0.284 & 0.557 & 0.794 \\
RND & 0.875 & 0.326 & 0.657 & 0.763 \\
PCA-kmeans & 0.710 & 0.282 & 0.744 & 0.651 \\
SAFE & 0.821 & 0.899 & 0.763 & 0.811 \\
\midrule
Current latent & 0.829 & 0.843 & 0.833 & 0.862 \\
Direct-linear & 0.875 & 0.963 & 0.833 & 0.898 \\
Single view & 0.930 & 0.925 & 0.907 & 0.890 \\
\textbf{ContactGuard} & 0.960 & 0.966 & 0.930 & 0.904 \\
\bottomrule
\end{tabular}
\end{minipage}
\end{table}

\textbf{Models and baselines:}
Our comparisons isolate the value of latent imagination and the relationship to established runtime failure detectors. \textbf{Ours} is the multi-view ContactGuard predictor that mean-pools the three camera latents before action-conditioned rollout. \textbf{Single view} is the same world model, including the same fixed projection, trained on the same unlabelled trajectories with only the middle camera, isolating the value of multi-view fusion. \textbf{Direct-linear} bypasses rollout and predicts failure directly from the anchor latent $z_t$ and planned action chunk using the same frozen multi-view encoder. \textbf{Current latent} is a matched imagination ablation: it uses the same encoder, labels, probe family, data pools, and train/validation protocol as ContactGuard, but replaces the predicted future latent with the current latent $z_t$. We also report \textbf{PCA-kmeans}, \textbf{FAIL-Detect}~\citep{xu2025faildetect}, \textbf{RND}~\citep{burda2019rnd}, and \textbf{SAFE}~\citep{gu2025safe} at the same gripper-closure trigger. FAIL-Detect and RND provide success-only uncertainty/novelty references, whereas SAFE provides an external supervised failure-detection reference. All world-model variants share the same frozen encoder, predictor backbone, and training objective. Appendix~\ref{app:direct_mlp} reports a nonlinear \emph{Direct} probe on the same $(z_t,a)$ inputs; it does not close the gap to the rollout.

\textbf{Forecasting and probing:}
At evaluation, the monitor triggers when gripper closure is expected within $k_{\mathrm{pre}}{=}15$ frames. We anchor $0.5$\,s before closure and roll out the world model for $K{=}30$ steps, reaching $0.5$\,s after closure at $30$\,Hz. A frozen $\ell_2$-regularized, class-balanced logistic-regression probe maps $\hat z_{t+K}$ to $P(\mathrm{fail})$. The probe is trained on the train split, selected on validation, and fixed for all test and real-robot results. Since the anchor precedes closure, the probe cannot directly observe the grasp outcome; any post-contact evidence must therefore be supplied through the predicted future latent.

\textbf{Closed-loop real-robot evaluation:}
We deploy ContactGuard with ACT using the same trigger, forecast horizon, frozen probe, and per-task threshold $\tau$. For evaluation, when the monitor would abort ($P(\mathrm{fail}){>}\tau$), the robot pauses $5$\,s with the gripper open and then resumes the remaining chunk to record the would-be outcome; unflagged trials run to completion. Since the pause precedes contact, the resumed attempt is a controlled counterfactual proxy for the abort decision. We also log all pre-contact observations and planned chunks from the live trials for offline replay for the external baselines and diagnostic ablations on identical inputs. Full operating points are in Appendix~\ref{app:full_tables}.

\begin{figure}[t]
\centering
\begin{subfigure}[t]{0.24\linewidth}
    \centering
    \includegraphics[width=\linewidth]{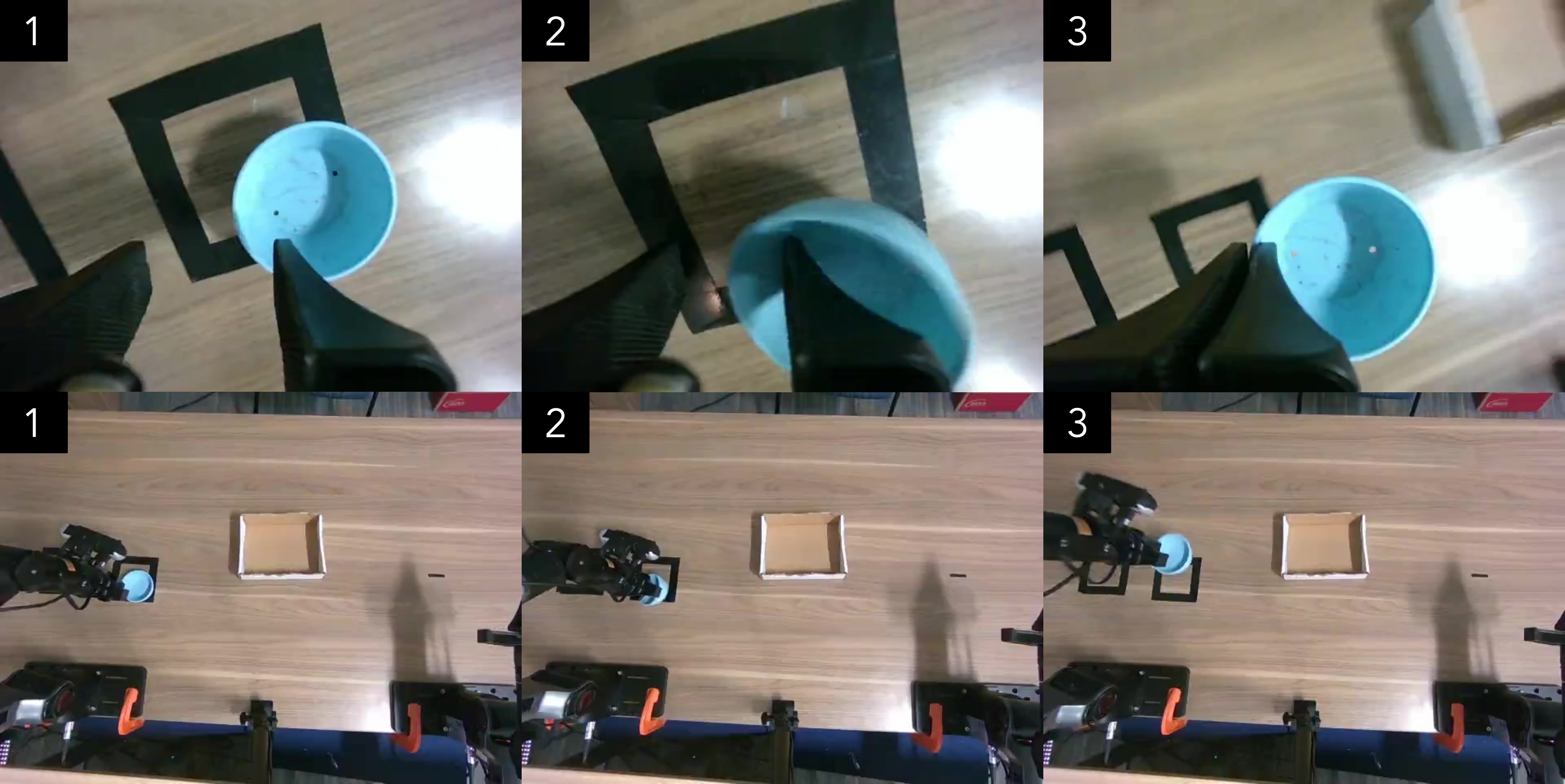}
    \caption{Pick-and-place (cup)}
\end{subfigure}
\hfill
\begin{subfigure}[t]{0.24\linewidth}
    \centering
    \includegraphics[width=\linewidth]{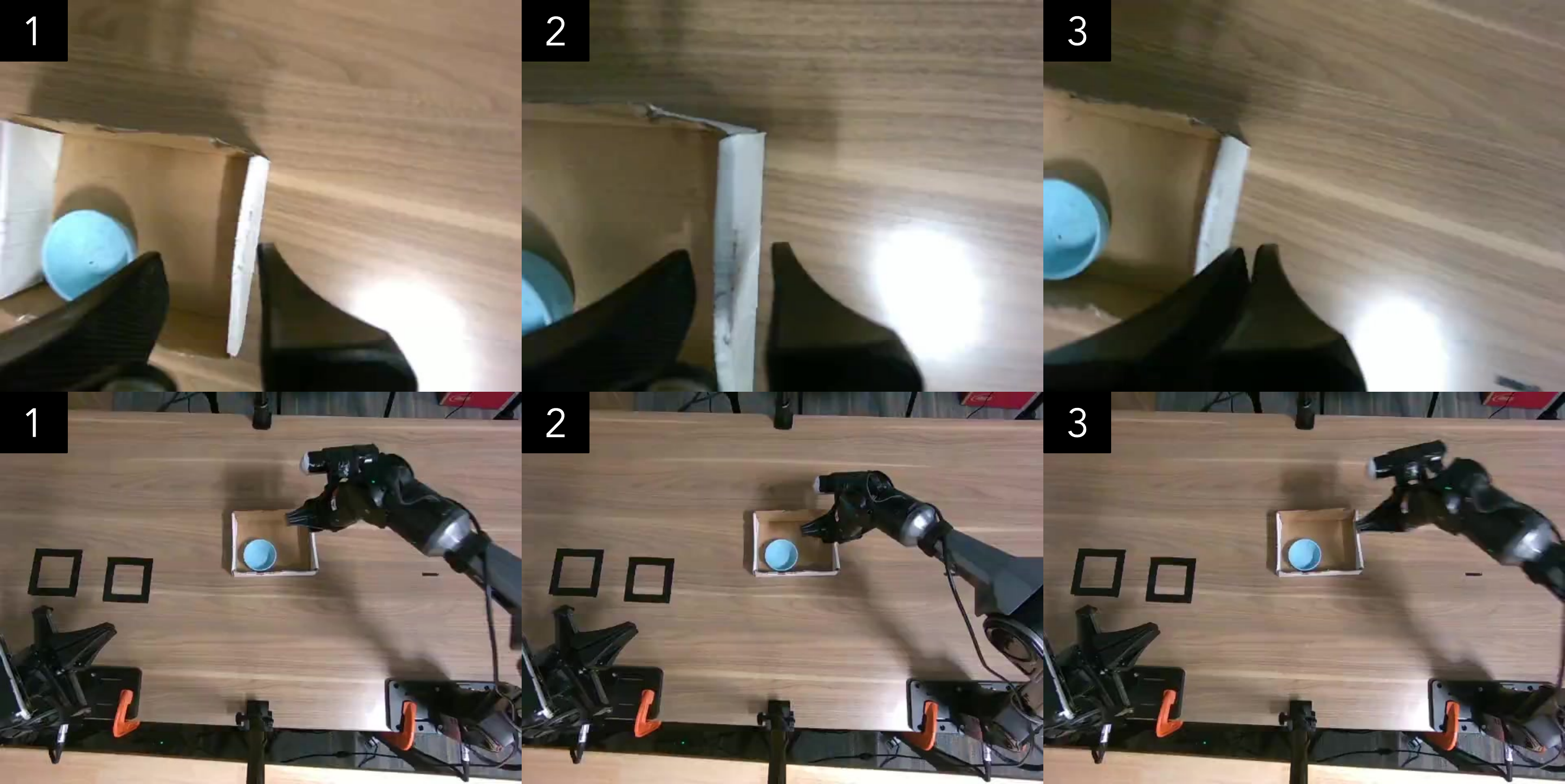}
    \caption{Pick-and-place (box)}
\end{subfigure}
\hfill
\begin{subfigure}[t]{0.24\linewidth}
    \centering
    \includegraphics[width=\linewidth]{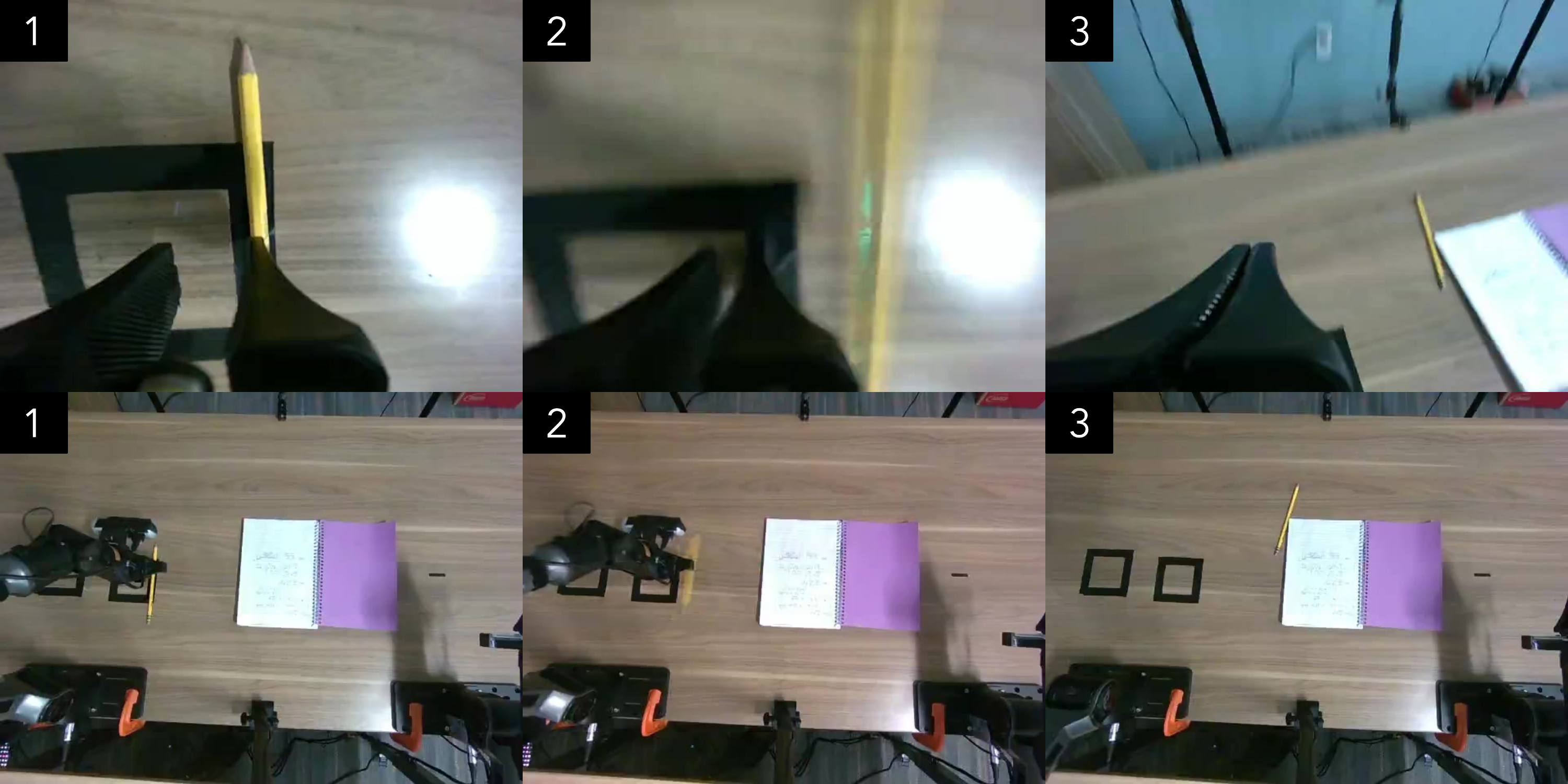}
    \caption{Pencil-and-notebook}
\end{subfigure}
\hfill
\begin{subfigure}[t]{0.24\linewidth}
    \centering
    \includegraphics[width=\linewidth]{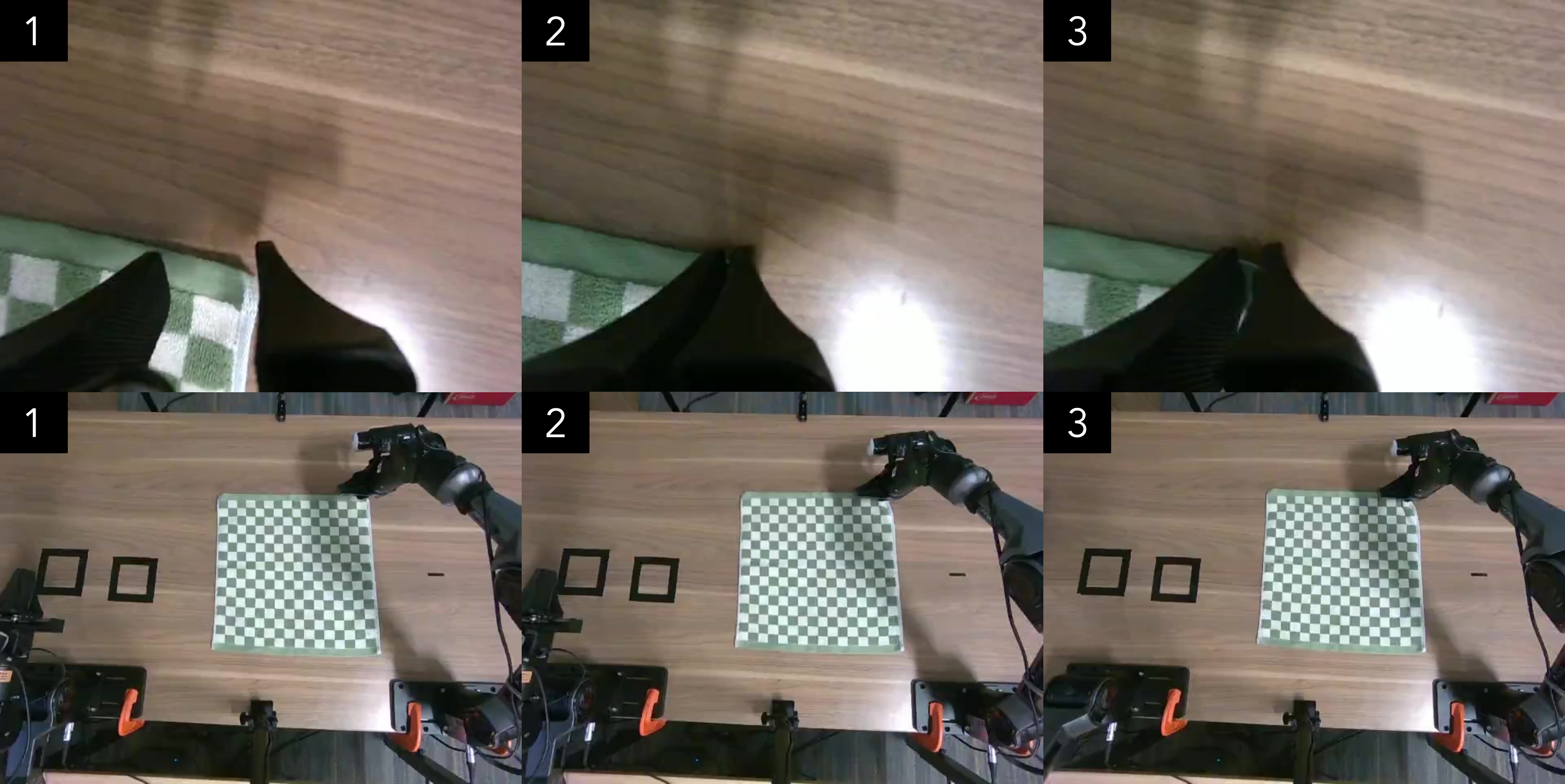}
    \caption{Towel-fold}
\end{subfigure}
\caption{Qualitative examples on the four tasks. Each panel shows a live rollout that our monitor classifies correctly.}
\label{fig:tasks}
\end{figure}

\begin{table}[t]
\centering
\begin{minipage}[t]{0.44\textwidth}
\vspace{0pt}
\small
\caption{Action perturbations on successful rollouts of Table~\ref{tab:realrobot}, with observations fixed. $\Delta p$: change in failure probability under the frozen probe. Transplant parentheses show the fraction with $\Delta p{>}0$. The action-free variant gives $\Delta p \equiv 0$ for all perturbations.}
\label{tab:ablations}
\end{minipage}\hfill
\begin{minipage}[t]{0.28\textwidth}
\vspace{0pt}
\small
\caption{Proprioceptive ablation: held-out AUC of multi-view ContactGuard with vs.\ without 28-d robot state on Table~\ref{tab:realrobot} attempts ($K{=}30$). Best in bold.}
\label{tab:proprio}
\end{minipage}\hfill
\begin{minipage}[t]{0.26\textwidth}
\vspace{0pt}
\small
\caption{RTX~5090 latency (FP32, batch 1, 200 timed passes, idle GPU). \emph{Full}: encode\,+\,rollout\,+\,classifier; \emph{Rollout}: rollout only.}
\label{tab:inference_cost}
\end{minipage}
\par\medskip
\begin{minipage}[b]{0.44\textwidth}
\centering
\small
\setlength{\tabcolsep}{3pt}
\begin{tabular}[b]{l c c c}
\toprule
Task & \twoline{Transplant}{$\Delta p$} & \twoline{Shift}{$\Delta p$} & \twoline{Hold}{$\Delta p$} \\
\midrule
Cup    & $+0.29$ (0.84) & $-0.01$ & $+0.12$ \\
Box    & $+0.54$ (0.79) & $+0.02$ & $+0.87$ \\
Pencil    & $+0.38$ (0.75) & $+0.00$ & $+0.02$ \\
Towel    & $+0.20$ (0.64) & $-0.05$ & $+0.11$ \\
\bottomrule
\end{tabular}
\end{minipage}\hfill
\begin{minipage}[b]{0.28\textwidth}
\centering
\small
\setlength{\tabcolsep}{3pt}
\begin{tabular}[b]{l c c}
\toprule
Task & \twoline{w/}{state} & \twoline{img}{only} \\
\midrule
Cup & 0.893 & \textbf{0.960} \\
Box & 0.906 & \textbf{0.966} \\
Pencil & 0.821 & \textbf{0.930} \\
Towel & 0.669 & \textbf{0.904} \\
\bottomrule
\end{tabular}
\end{minipage}\hfill
\begin{minipage}[b]{0.26\textwidth}
\centering
\small
\setlength{\tabcolsep}{3pt}
\begin{tabular}[b]{@{}c c c c@{}}
\toprule
$K$ & \twoline{Horizon}{(s)} & \twoline{Rollout}{(ms)} & \twoline{Full}{(ms)} \\
\midrule
5  & 0.17 &  3.08 &  9.02 \\
15 & 0.50 &  9.26 & 13.24 \\
25 & 0.83 & 14.31 & 18.71 \\
30 & 1.00 & 17.20 & 21.48 \\
\bottomrule
\end{tabular}
\end{minipage}
\end{table}

\subsection{Closed-Loop Grasp-Outcome Prediction}
\label{sec:realrobot}

Table~\ref{tab:realrobot} reports closed-loop metrics from live rollouts; the world model, probe, and per-task threshold are frozen beforehand. All subsequent diagnostics are computed offline on the logged rollout data.

\textbf{Multi-view vs.\ single-view (Table~\ref{tab:realrobot}):} Multi-view fusion improves both balanced accuracy and AUC over the \textbf{Single view} baseline across all four tasks.
The gain in AUC is largest on Box, and smallest on Towel. Towel yields our model's lowest absolute AUC, reflecting the difficulty of perceiving the gripper--cloth contact region even with multiple views.

\textbf{Direct prediction without imagination:}
The \emph{Direct-linear} baseline receives both the same frozen anchor latent and the planned action chunk, but must learn the outcome mapping directly from the small labelled probe set. It trails ContactGuard in AUC on every live task, with the largest gaps on Cup and Pencil; on Box the two AUCs are nearly tied, but Direct-linear incurs a higher false-abort rate at the frozen threshold. Together with the matched \emph{Current latent} comparison in Table~\ref{tab:external_baselines}, this shows that the gain does not come merely from the current representation or planned actions: rolling those actions through the pretrained latent dynamics produces a more useful failure representation.

\textbf{External failure detectors (Table~\ref{tab:external_baselines}):}
Evaluated on the saved real-robot rollouts, ContactGuard achieves the highest AUC across all four tasks. The matched \emph{Current latent} ablation uses the same encoder, labels, probe family, data pools, and train/validation protocol, differing only in whether the probe reads the current latent or the action-conditioned imagined future. Its consistently lower performance supports the benefit of imagined consequences beyond the current observation alone. ContactGuard also outperforms the tested external detectors, while FAIL-Detect and RND are less consistently aligned with failure risk in this pre-contact setting.

\textbf{Per-task abort behaviour (Table~\ref{tab:realrobot}):}
The four tasks abort in distinct ways.
Cup behaves well, with realised $P(\text{fail})$ scores providing strong separation between true failures and true successes.
Box trades precision for coverage: it catches all failures, but has the highest false-abort rate among the four tasks.
Towel is asymmetric in the opposite direction: it rarely aborts a successful grasp, but a non-trivial fraction of failures score inside the success range and slip through, giving it the weakest recall.
Pencil falls in between, with strong balanced accuracy but lower recall than Box or Cup.

\subsection{Offline Diagnostics: What Information Does the Monitor Use?}
\label{sec:ablations_sec}

We now ask \emph{why} the monitor works: does proprioceptive state help, and does the signal come from the rolled-out latent and aligned actions? The action perturbations are diagnostic offline interventions, so Table~\ref{tab:ablations} reports open-loop replay at $K{=}30$ on logged real-robot inputs.

\textbf{Proprioceptive input (Table~\ref{tab:proprio}):}
We compare the multi-view ContactGuard world model with vs.\ without the 28-d robot state (joint positions and efforts). In the state variant, each frame's state is embedded by a small MLP and fused with the image latent through concatenation and a linear layer; like the encoder, this fusion is kept fixed at its initialization. The two variants use the same latent dimensionality, training recipe, data, held-out attempts, and \(K{=}30\) probe protocol. Counter to the usual intuition, dropping the state improves test AUC on all four tasks. This suggests that under our simple fixed-fusion design and small-data regime, proprioceptive state does not provide a robust complementary signal and may instead introduce task-specific shortcuts. We do not claim proprioception is harmful under richer fusion or larger datasets.

\textbf{Action and latent information (Table~\ref{tab:ablations}):}
We next test whether ContactGuard scores the consequence of the \emph{specific pending action}, rather than merely recognizing a risky pre-contact state. Replacing the predicted future latent with the current anchor latent consistently reduces AUC across all four tasks (Table~\ref{tab:external_baselines}), indicating that the rollout adds information beyond the pre-contact observation.

We then hold the visual observation fixed and perturb only the action input. \emph{Transplant} replaces the planned action chunk of a successful attempt with one randomly sampled from a failed attempt of the same task; \emph{Shift} advances the original chunk by five frames and pads the tail with its final action; and \emph{Hold} repeats the first action for the full rollout horizon. Transplant generally increases the predicted failure probability, whereas Shift and Hold produce task-dependent responses; an action-free variant is invariant to all three perturbations. These results show that ContactGuard is sensitive to the specific pending action rather than relying only on static visual risk.

\textbf{Runtime analysis:} Online monitoring must return a verdict before the gripper reaches the candidate grasp frame. We benchmark the monitor, with its frozen DINOv2-S encoder and 4-layer transformer predictor, on a single NVIDIA RTX~5090 in FP32 at the deployment input shape; per-horizon costs are reported in Table~\ref{tab:inference_cost}. History encoding is a one-shot cost shared across rollout horizons, while rollout latency scales nearly linearly with $K$ because the predictor autoregressively unrolls over cached context tokens. Peak CUDA memory is dominated by encoder activations and remains essentially independent of $K$, since each rolled-out latent is only $D$-dimensional. At the deployed horizon, the full encode--rollout--probe pass takes about $22$\,ms, within the $33$\,ms control period, so the monitor does not bottleneck control.


\section{Conclusions, Limitations, and Future Work}
\label{sec:conclusion}

We presented \emph{ContactGuard}, a real-time, policy-decoupled predictive verifier for pre-contact manipulation. An independently trained action-conditioned latent world model evaluates the concrete action chunk proposed by an otherwise unchanged visuomotor policy and scores its predicted post-contact consequence with a lightweight failure probe. Across four real-world grasp settings, ContactGuard outperforms matched current-state monitoring and the tested external runtime failure detectors. Holding the observation fixed while replacing only the proposed action also changes the predicted failure probability substantially, confirming that the monitor responds to the pending action rather than only to static visual risk. These results support action-conditioned latent imagination as a practical substrate for vetoing likely failures before scene-changing contact.

\subsection{Limitations}
ContactGuard prevents failures by abstaining, but does not recover from them or complete the task after an abort. Post-abort task completion requires an external recovery module, which we leave for future work. ContactGuard targets imminent contact events whose outcome is determined within the next action chunk; extending it to longer-horizon skills would require hierarchical or repeated event-level monitoring.

\textbf{Future work:}
Future work can close the loop after an abort by selecting a new action chunk, replanning from the preserved scene, or coupling the monitor with multi-sample policies such as Diffusion~\citep{chi_diffusion_2024}, Flow Matching~\citep{lipman2023flow}, or Streaming~\cite {safe_post_training} policies. 



\clearpage

\bibliography{example}  

\clearpage
\appendix

\section{Implementation Details}
\label{app:impl}

\paragraph{Encoder and latent dimensions.}
Camera images are resized to $224{\times}224$ before encoding.
Each view is encoded by a shared, frozen DINOv2-S (ViT-S/14, \texttt{vit\_small\_patch14\_dinov2.lvd142m}~\citep{oquab_dinov2_2024}); its 384-d CLS embeddings are mean-pooled across cameras and mapped through a fixed linear projection, giving $z_t \in \mathbb{R}^{384}$ (21.8 M frozen parameters). The single-view model applies the same fixed projection to the middle-camera CLS embedding.

\paragraph{Action embedder.}
The per-step action embedder is a $1{\times}1$ temporal convolution followed by a two-layer MLP that maps each $\mathbb{R}^d$ action to a 384-dimensional embedding.

\paragraph{Predictor.}
The autoregressive predictor stacks 4 AdaLN-zero conditioned causal Transformer blocks of width 384 with 8 attention heads of dimension 64, a feed-forward hidden size of 1536, and dropout 0.1. The prediction head is a 2-layer MLP with hidden size $4D = 1536$; during rollout each predicted latent is passed through this head before being fed back into the context. The trainable part of the model (predictor, action embedder, and head) has 13.2\,M parameters; the full monitor has 35.0\,M.

\paragraph{Horizons.}
We use a training context window $C{=}3$. Each training window supplies real latents at the offsets $\mathcal{H}=\{1,2,3,5,8,12,18,24,30\}$ beyond the context: the first three serve as teacher-forced one-step targets ($m{=}3$ in Eq.~\eqref{eq:loss}), and all nine supervise the free-running 30-step rollout. At deployment and offline probe evaluation the rollout horizon is $K{=}30$, with anchor offset $k_{\mathrm{pre}}{=}15$ frames before the planned closure event (0.5\,s at 30\,Hz), so the probe reads out $\hat z_{t+K}$ at $T_g + 15$ frames (0.5\,s post-closure).

\paragraph{Training objective and optimizer.}
The loss is the whitened regression of Eq.~\eqref{eq:loss}, with the per-dimension variances $v_d$ in $\Sigma$ estimated once before training from 7{,}200 encoder latents, so that the loss divides each latent dimension by its standard deviation $\sqrt{v_d}$; rollout targets are stop-gradient. No SIGReg or other anti-collapse term is used, since the encoder is frozen. The predictor is trained from scratch for 30 epochs with AdamW (learning rate $10^{-4}$, weight decay $10^{-3}$), a cosine learning-rate schedule, batch size 40, gradient clipping at 1.0, and mixed-precision training. The single-view ablation uses the identical configuration with only the middle camera as input.

\paragraph{Probe.}
The logistic regression probe of Eq.~\eqref{eq:probe_loss} uses $\ell_2$ regularization with $\rho{=}1$ and class-balanced reweighting. Standardization and probe fitting use the training split, while model selection and the per-task threshold use validation data. Final evaluation uses $N{=}50$ separately collected live real-robot rollouts per task, excluded from all training and tuning. The same rollouts are used for all offline baseline and diagnostic comparisons.

\begin{table}[htbp]
\centering
\small
\caption{Probe capacity on real-robot rollouts ($K{=}30$). ROC AUC and balanced accuracy for the linear variants and nonlinear \emph{Direct} probes. Nonlinear capacity does not consistently improve Direct-linear, and all Direct variants remain below ContactGuard.}
\label{tab:supp_fullrollout}
\begin{tabular}{l l c c}
\toprule
Task (split) & Model & test AUC & test bal-acc \\
\midrule
\multirow{5}{*}{Cup}
 & Ours      & \textbf{0.960} & \textbf{0.900} \\
 & Single view      & 0.930 & 0.860 \\
 & Direct-linear & 0.875 & 0.760 \\
 & Direct-MLP small & 0.866 & 0.671 \\
 & Direct-MLP large & 0.870 & 0.540 \\
\midrule
\multirow{5}{*}{Box}
 & Ours      & \textbf{0.966} & \textbf{0.875} \\
 & Single view      & 0.925 & 0.852 \\
 & Direct-linear & 0.963 & 0.781 \\
 & Direct-MLP small & 0.943 & 0.602 \\
 & Direct-MLP large & 0.815 & 0.626 \\
\midrule
\multirow{5}{*}{Pencil}
 & Ours      & \textbf{0.930} & \textbf{0.878} \\
 & Single view      & 0.907 & 0.823 \\
 & Direct-linear & 0.833 & 0.769 \\
 & Direct-MLP small & 0.844 & 0.591 \\
 & Direct-MLP large & 0.799 & 0.483 \\
\midrule
\multirow{5}{*}{Towel}
 & Ours      & \textbf{0.904} & \textbf{0.860} \\
 & Single view      & 0.890 & 0.800 \\
 & Direct-linear & 0.898 & 0.740 \\
 & Direct-MLP small & 0.787 & 0.605 \\
 & Direct-MLP large & 0.798 & 0.758 \\
\bottomrule
\end{tabular}
\end{table}

\section{Nonlinear \emph{Direct} Variants}
\label{app:direct_mlp}

We also evaluated nonlinear \emph{Direct} probes using the same inputs as Direct-linear: the frozen anchor latent $z_t$ concatenated with the planned action chunk $a_{t:t+K-1}$. We swept 2- and 3-layer MLPs over hidden width, dropout, and weight decay, with early stopping on validation AUC. As shown in Table~\ref{tab:supp_fullrollout}, increasing probe capacity does not consistently improve the direct baseline, and all nonlinear Direct variants remain below ContactGuard. This supports our small-data design choice: rather than relying on a more expressive probe to learn the outcome mapping directly from $(z_t,a)$, ContactGuard uses the pretrained action-conditioned rollout to produce a future latent that is effective with a low-capacity linear readout.

\section{Per-Task Operating Points}
\label{app:full_tables}

Table~\ref{tab:abort_rate_supp} expands the results of Table~\ref{tab:realrobot} into the full operating point at the frozen threshold, including class counts, thresholds, confusion matrices, false-abort rates, and classifier metrics.

\begin{table}[htbp]
\centering
\small
\caption{Per-task operating points on real-robot rollouts. Class counts ($N_{+}/N_{-}$), the frozen threshold $\tau$ (selected by Youden-$J$ on the validation split), the confusion matrix (TP/FP/TN/FN), recall, false-abort rate, balanced accuracy, precision, and ROC AUC. The positive class is an imminent failed grasp, so TP is a correctly vetoed failure and FP a false abort.}
\label{tab:abort_rate_supp}
\setlength{\tabcolsep}{2pt}
\begin{tabular}{l l c c c c c c c c}
\toprule
Task & Method & $N_{+}/N_{-}$ & $\tau$ & TP/FP/TN/FN & Recall $\uparrow$ & FAR $\downarrow$ & bal-acc $\uparrow$ & Precision $\uparrow$ & AUC $\uparrow$ \\
\midrule
\multirow{3}{*}{Cup}
 & Ours & 25/25 & 0.487 & 22/2/23/3  & \textbf{0.880} & \textbf{0.080} & \textbf{0.900} & \textbf{0.917} & \textbf{0.960} \\
 & Single view & 25/25 & 0.194 & 21/3/22/4  & 0.840 & 0.120 & 0.860 & 0.875 & 0.930 \\
 & Direct-linear & 25/25 & 0.301 & 17/4/21/8  & 0.680 & 0.160 & 0.760 & 0.810 & 0.875 \\
\midrule
\multirow{3}{*}{Box}
 & Ours & 22/28 & 0.022 & 22/7/21/0  & \textbf{1.000} & \textbf{0.250} & \textbf{0.875} & \textbf{0.759} & \textbf{0.966} \\
 & Single view & 22/28 & 0.087 & 21/7/21/1  & 0.955 & \textbf{0.250} & 0.852 & 0.750 & 0.925 \\
 & Direct-linear & 22/28 & 0.018 & 21/11/17/1 & 0.955 & 0.393 & 0.781 & 0.656 & 0.963 \\
\midrule
\multirow{3}{*}{Pencil}
 & Ours & 22/28 & 0.519 & 19/3/25/3  & \textbf{0.864} & 0.107 & \textbf{0.878} & 0.864 & \textbf{0.930} \\
 & Single view & 22/28 & 0.859 & 15/1/27/7  & 0.682 & \textbf{0.036} & 0.823 & \textbf{0.938} & 0.907 \\
 & Direct-linear & 22/28 & 0.195 & 15/4/24/7  & 0.682 & 0.143 & 0.769 & 0.789 & 0.833 \\
\midrule
\multirow{3}{*}{Towel}
 & Ours & 25/25 & 0.613 & 19/1/24/6  & \textbf{0.760} & \textbf{0.040} & \textbf{0.860} & \textbf{0.950} & \textbf{0.904} \\
 & Single view & 25/25 & 0.193 & 16/1/24/9  & 0.640 & \textbf{0.040} & 0.800 & 0.941 & 0.890 \\
 & Direct-linear & 25/25 & 0.051 & 13/1/24/12  & 0.520 & \textbf{0.040} & 0.740 & 0.929 & 0.898 \\
\bottomrule
\end{tabular}
\end{table}

\end{document}